\documentclass[12pt]{article}
\usepackage[papersize={8.5in,11in},margin=1in]{geometry}

\usepackage{sectsty}
\sectionfont{\fontsize{13}{15}\selectfont}
\subsectionfont{\fontsize{12}{13}\selectfont}

\usepackage{float}
\usepackage{mathtools}
\usepackage{graphicx}
\usepackage{url}
\usepackage{multirow}
\usepackage{makecell}
\usepackage[square,comma]{natbib} 
\usepackage{tabularx}
\usepackage{caption,subcaption}
\usepackage[export]{adjustbox}
\usepackage{amsmath}
\DeclareMathOperator*{\argmax}{argmax}

\usepackage{hyperref}
\hypersetup{
  pdfcreator = {},
  pdfproducer = {}
}
\usepackage[math]{iwona}
\usepackage{mathptmx}

\begin{document}

\begin{center}

{\Large \bf{Don't Classify, Translate:\\
Multi-Level E-Commerce Product Categorization \\
Via Machine Translation}
}\\
\vspace{2mm}
{\it Completed Research Paper}

\vspace{5mm}
\begin{tabular}{ccc}
{\bf Maggie Yundi Li $\,\,\,\,$ Stanley Kok} &  & {\bf Liling Tan} \\
School of Computing  &  & Rakuten Institute of Technology \\
National University of Singapore  &  & \texttt{liling.tan@rakuten.com} \\
\texttt{a0131278@comp.nus.edu.sg} & &  \\
\texttt{skok@comp.nus.edu.sg} & &  \\

\end{tabular}
\end{center}

\begin{center}
{\large Abstract}\\
\end{center}

\noindent
E-commerce platforms categorize their products into a multi-level taxonomy tree with thousands of leaf categories. Conventional methods for product categorization are typically based on machine learning {\em classification} algorithms. These algorithms take product information as input (e.g., titles and descriptions) to classify a product into a leaf category. In this paper, we propose a new paradigm based on {\em machine translation}. In our approach, we translate a product's natural language description into a sequence of tokens representing a root-to-leaf path in a product taxonomy. In our experiments on two large real-world datasets, we show that our approach achieves better predictive accuracy than a state-of-the-art classification system for product categorization. In addition, we demonstrate that our machine translation models can propose meaningful new paths between previously unconnected nodes in a taxonomy tree, thereby transforming the taxonomy into a directed acyclic graph (DAG). We discuss how the resultant taxonomy DAG promotes user-friendly navigation, and how it is more adaptable to new products.
%

\vspace{1mm}


\section{Introduction}
\label{sect:intro}

\begin{figure}[t]
\centering
\includegraphics[width=\textwidth]{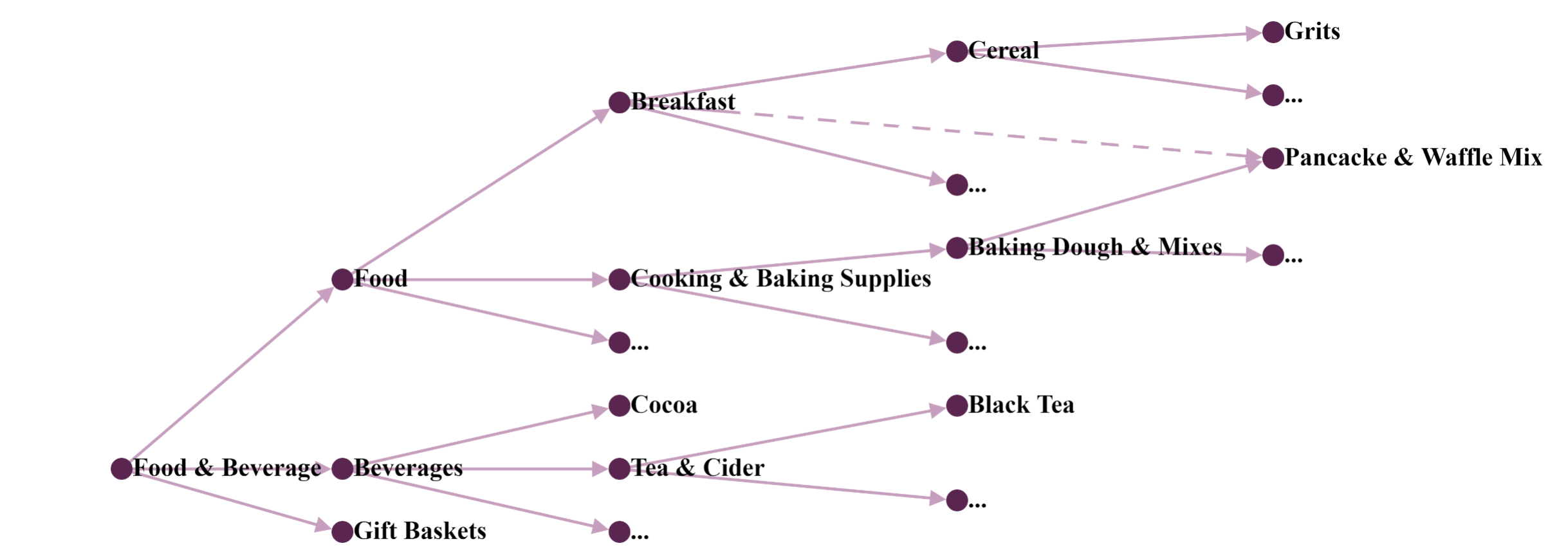}
\caption{Example of a product catalog organized in a tree structure (solid lines and circles). Individual products are added to leaf nodes (e.g., {\tt Grits}). The addition of the dotted edge turns the tree into a directed acyclic graph (DAG), and there are now more than one root-to-leaf paths to the leaf node {\tt Pancake \& Waffle Mix}.}
\label{fig:catalog-eg}
\end{figure}

Product catalogs are critical to e-commerce platforms such as Alibaba, Amazon, Rakuten, and Shopee. These catalogs typically categorize millions of products into a taxonomy tree three to ten levels deep with thousands of leaf nodes~\citep{Shen2012,mcauley&al15} (Figure~\ref{fig:catalog-eg}), and are continually updated with millions of new products per month from thousands of merchants. Correctly categorizing a new product into the taxonomy is fundamental to many business operations, such as enforcing category-specific listing and censorship policies, extracting and presenting relevant product attributes, and determining appropriate handling and shipping fees. Further, the accuracy and coherency of the taxonomy play important roles in customer-facing services such as product search and  recommendation, and user browsing and navigation of the catalog. Clearly, given the large scale of the product taxonomy, manually categorizing a new product into it is both unscalable and error-prone, and we need automated algorithms for doing so. 

To date, algorithms for product categorization have largely formulated the problem as a standard machine learning {\em classification} task, which takes the textual description of a product as input (e.g., {\it ``Mix Pancake Waffle 24 OZ -Pack of 6"}) and outputs a leaf node that is the product's most likely category. Because the taxonomy is a tree and each leaf node uniquely defines a path from root to leaf, these algorithms are effectively outputting an {\em existing} root-to-leaf path. Modulo the addition of the product to the leaf, these algorithms do not alter the taxonomy's tree structure.

In contrast to classification-based approaches, we map the problem of product categorization to the task of machine translation (MT). An MT system takes text in one language as input (traditionally denoted as $f$) and outputs its translation as a sequence of words in another language (denoted as $e$). The input $f$ maps to the textual description of a product, and the output $e$ maps to the sequence of categories and sub-categories in a root-to-leaf path (e.g., {\tt  Baking Supplies \ $\rightarrow$ \ Flour \& Dough \ $\rightarrow$ \ Pancake \& Waffle Mixes}). By framing product categorization as an MT problem, our approach offers several operational and technical advantages over previous algorithms.

First, large e-commerce companies typically operate their sites globally in a variety of languages (e.g., {\it www.rakuten.com} in English and {\it www.rakuten.co.jp} in Japanese), and have invested heavily in their machine translation capabilities. By utilizing these existing MT systems for the task of product categorization (rather than developing a new disparate system), we are reducing the technical debt that these companies incur. They have fewer algorithms to be cognizant of, fewer systems to develop, less bugs to fix, and consequently lower maintenance cost.    

Second, machine translation systems, through the use of deep learning~\citep{goodfellow&al16}, have improved their accuracy by leaps and bounds in recent years~\citep{kalchbrenner&blunsom13,sutskever&al14,bahdanau&al15}, even to the extent of achieving human parity on some language pairs~\citep{hassan&al18}. By mapping the problem of product categorization to one of machine translation, we bring the best of MT technology to bear on the problem of product categorization in a cost-effective manner. In Section~\ref{sect:experiments}, we provide empirical results demonstrating that our MT approach outperforms state-of-the-art systems. 

Third, machine translation systems are by nature resilient to the vagaries and noise present in language, and thus are robust to errors in a product's textual description and the varieties of ways in which a product can be specified (e.g., {\it ``Mix Pancake Waffle 24 OZ -Pack of 6"} and {\it ``Packet of six; waffle pancake mix; 24 ounces"} refer to the same product). This makes MT systems ideal for dealing with the uncertainties inherent in a product's natural language description.

Fourth, our MT approach not only outputs pre-existing root-to-leaf paths in a taxonomy tree, it also produces novel root-to-leaf paths that do not exist in the taxonomy. These novel paths transform the structure of a product catalog from a tree to a directed acyclic graph (DAG). This is a powerful transformation, offering potentially multiple root-to-leaf paths to a single product (rather than just one path as in previous systems). This better conforms to psychological findings that humans tend to view an object in multiple ways~\citep{heit&rubinstein94,ross&murphy99,shafto&coley03,shafto&al05}. For example, a waffle is regarded as being primarily carbohydrates because it is made of flour; however, it is often also regarded as a breakfast food. The different ways of thinking about a waffle underline the different ways of thinking about food: as a system of taxonomic categories like flour and sugar, or as situational categories like breakfast foods and dinner fare. Likewise, in other product domains, items have different properties, and more than one system of categories are required to fully represent these properties. By creating multiple root-to-leaf paths, our system better caters  to human intuition than previous systems, and can potentially improve customer-facing applications such as user product navigation. For example, by having both of the following paths in a product DAG, a user who predominantly views a waffle mix as baking supplies, and a user who views it as a breakfast food can both expeditiously navigate to what they need.
\begin{itemize}
\itemsep0em
\item {\tt  Baking Supplies \ $\rightarrow$ \  Flour \& Dough \ $\rightarrow$ \ Pancake \& Waffle Mixes}
\item {\tt Breakfast Foods \ $\rightarrow$ \ Pancake \& Waffle Mixes}
\end{itemize}

To our knowledge, we are the first to apply a machine translation approach to the problem of e-commerce product categorization.

Next, we briefly review related work (Section~\ref{sect:related}). We then describe state-of-the-art machine translation systems, and how we use them for product categorization (Section~\ref{sect:mt}). Next, we describe our datasets, experimental methodology, comparison systems, and empirical results (Section~\ref{sect:experiments}). Then, we provide a qualitative analysis of our results (Section~\ref{sect:analysis}). Finally, we conclude with future work (Section~\ref{sect:conclusion}).

\section{Related Work}
\label{sect:related}

The vast majority of product categorization systems is based on machine learning {\em classification} algorithms. These systems can be dichotomized into (a) those that classify a product in a single step into one of thousands of leaf nodes in a taxonomy tree, and (b) those that classify the product step-wise, first into higher-level categories, and then into lower-level subcategories. This dichotomy is due to the inherent skewness (long tail phenomenon~\citep{anderson06}) that is typical of e-commerce products -- a large proportion of products are distributed over a small number of categories (leaf nodes) with the remaining fraction of products sprinkled over a large number of remaining categories. This imbalance of category sizes poses a challenge to classification algorithms, which generally require the sizes of categories to be approximately balanced. To circumvent this problem, the step-wise approach first performs classification across top-level categories, each of which aggregates products in its lower-level subcategories to ameliorate the data imbalance problem. After assigning a product to a top-level category, the step-wise approach repeats the process and performs classification across the sub-categories. Because these sub-categories belong to the same top-level category, they are likely to belong to the same class of products, and hence have less imbalance in their sizes. However, the step-wise approach suffers from two shortcomings: (a) errors from classifiers at previous steps get propagated to classifiers at subsequent steps with no chance of recovery, and (b) the number of classifiers grows exponentially with every step. Unlike the step-wise systems, the single-step approach does not have these drawbacks, but must contend with the category imbalance problem at its full severity at the leaf nodes.

A variety of single-step classifiers have been used for product categorization. \citeauthor{yu&al13} (\citeyear{yu&al13}) explore a gamut of word-level features (e.g., n-grams), and use a support vector machine (SVM;~\cite{cortes1995support}) as their classification algorithm. \citeauthor{chen&warren13} (\citeyear{chen&warren13}) sensitize the objective function of an SVM to the average revenue loss of erroneous product classifications, thereby trading high revenue-loss errors for low revenue-loss ones. \citeauthor{sun&al14} (\citeyear{sun&al14}) use simple classifiers (e.g., naive Bayes, k-nearest neighbors, and perceptron), and recruit manual labor via crowdsourcing to flag their errors. \citeauthor{Kozareva15} (\citeyear{Kozareva15}) uses a variety of features (e.g., n-grams, latent Dirichlet allocation topics~\citep{blei2003latent}, and word2vec embeddings~\citep{mikolov&al13}) in a multi-class algorithm.
Both \citeauthor{ha2016large} (\citeyear{ha2016large}) and \citeauthor{xia17} (\citeyear{xia17}) use deep learning to learn a compact vector representation of the attributes of a product (e.g., product title, merchant ID, and product image), and use the representation to classify the product. They differ in terms of the kinds of deep learning model used. The former uses recurrent neural networks~\citep{hochreiter&schmidhuber97} and the latter uses convolutional neural networks~\citep{lecun&al98}.

Several step-wise classifiers have also been used for product categorization. \citeauthor{Shen2012} (\citeyear{Shen2012}) use simple classifiers (e.g., naive Bayes and k-nearest neighbors) in the first step, then an SVM to assign a product to a leaf node in the second step.  \citeauthor{das2016large} (\citeyear{das2016large}) explore the use of gradient boosted trees~\citep{friedman00} and convolutional neural networks in each of three steps. However, they only evaluated the accuracy of their approach at the top two levels of a product taxonomy (a simpler problem because of the smaller number of categories at the top levels), and did not provide the accuracy at the leaf nodes. \citeauthor{cevahir2016large} (\citeyear{cevahir2016large}) use deep belief networks (DBNs;~\cite{hinton2006fast}) and k-nearest neighbors (KNNs) in a two-step approach. Because the number of models grows exponentially with the number of steps, a large number of models are trained (72) even though only two steps are involved. This large number of models makes it impractical to deploy their approach in a real-world production setting. They also use a single-step approach (termed \texttt{CUDeep}) consisting of one DBN and one KNN, and found that it is competitive against the 72-model, two-step approach. With only two models, their single-step approach trains faster and is feasible for real-world deployment. In our experiments in Section~\ref{sect:experiments}, we compare our machine translation (MT) approaches against \texttt{CUDeep}.

All of these classification systems assign a product into an existing leaf node (which is equivalent to a unique existing root-to-leaf path). Unlike them, our machine translation approach is able to create both existing root-to-leaf paths and novel non-existing paths for a product, thereby presenting a richer representation of a product to both downstream business operations and customer-facing applications. In addition, our MT approach outperforms classification algorithms in terms predictive accuracy (results in Section~\ref{sect:experiments}). 

\section{Machine Translation Systems}
\label{sect:mt}

A machine translation (MT) system takes as input a sentence ${\bf f} = f_1 f_2 \ldots f_m$ with $m$ tokens in a {\em source} language, and finds its best translation ${\bf e} = e_1 e_2 \ldots e_n$ with $n$ tokens in a {\em target} language (this is expressed mathematically as ${\bf e} = \argmax_{\bf e} P({\bf e}|{\bf f})$).

In the past, the predominant MT approach was phrase-based machine translation (PBMT), which is grounded in information theory and statistics.
Though moderately successful in its heyday, it has recently been eclipsed by neural machine translation (NMT) approaches that utilize deep learning~\citep{goodfellow&al16}. Deep learning contributes to NMT by incrementally building more sophisticated models of languages, and then linking them to models of the translations.

First, deep learning compresses the common 1-of-N representation of a word (i.e., an N-dimensional vector with a single 1 at the index corresponding to the word and 0's everywhere else) into a smaller continuous-valued feature vector (also known as an embedding) ~\citep{mikolov&al13}. Intuitively, this vector provides a distributed continuous representation of an input word, with the vector's continuous values varying gradually among similar words and differing greatly among dissimilar ones. Such vectors are then used to represent a probability distribution over the words they represent. The continuity in the vectors automatically smoothens the distribution and alleviates the data sparsity problem (this occurs when the vocabulary of a language is large, and too few occurrences of various words appear in a corpus).

\begin{figure}[t]
\begin{center}
\includegraphics[width=0.5\textwidth]{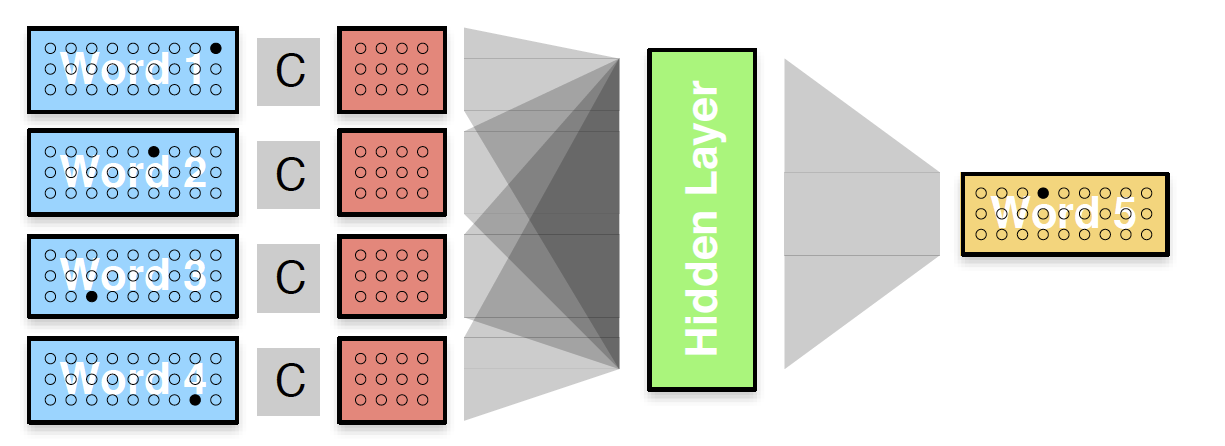}
\caption{A language model encoded by a feed forward neural network. Previous 4 words are represented as 1-of-N vectors (blue), compressed into continuous vectors (red) using the same matrix C for all words, passed through a hidden layer, and then used to predict the next word as a 1-of-N vector (yellow)  (Image from \citep{koehn17})}
\label{fig:fnn}
\end{center}
\end{figure}

Second, deep learning allows complex features to be learned automatically from text. Building upon the vector embeddings of words, we could connect these to another layer of vectors that are collectively termed {\em hidden} layers, and in turn, connect those to other hidden layers. As more hidden layers are added (one on top of another) to form a {\em feedforward neural network},  they can model more complex interactions and features among words in the input text. Feedforward neural networks can model a language by using the previous $n$ words in a sentence to predict the current word (Figure~\ref{fig:fnn}). This way a feedforward neural network encodes the probability distribution of a next word given its previous words as context.   

\begin{figure}[t]
\begin{center}
\includegraphics[width=0.45\textwidth]{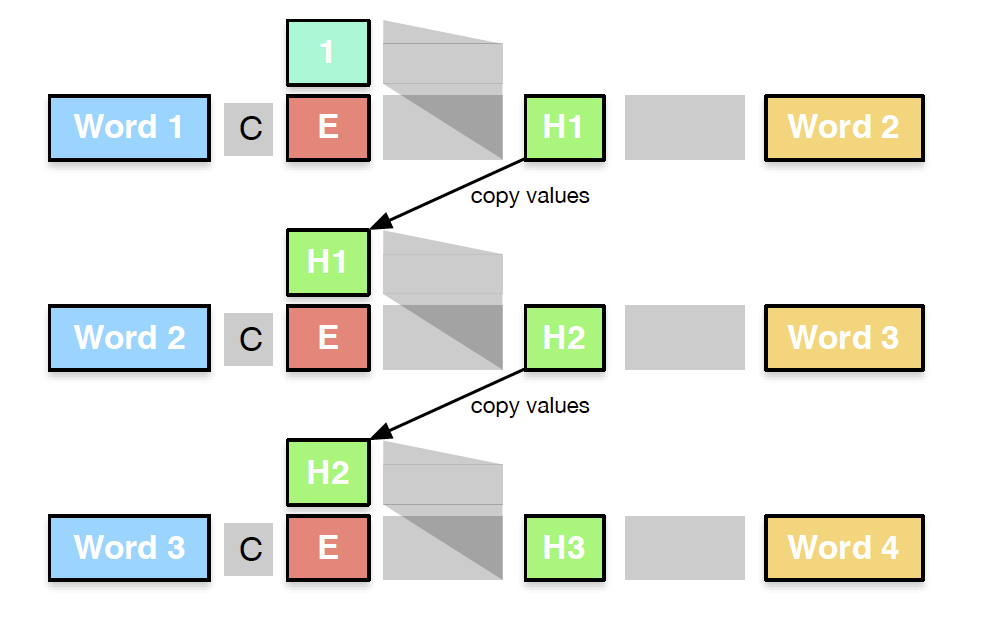}
\caption{A language model encoded as a recurrent neural network. After predicting Word 2 (yellow), we reuse the hidden layer H1 (green) together with the correct Word 2 (blue) to predict Word 3 (yellow).  (Image from \citep{koehn17})}
\label{fig:rnn}
\end{center}
\end{figure}

Third, more powerful language models can be built using recurrent neural networks (RNNs) \citep{hochreiter&schmidhuber97}. An RNN is similar to a feedforward neural network in having an input layer of words that is connected to a hidden layer, which in turn is connected to an output layer representing a probability distribution over words. It differs by linking the hidden layer back to itself with recurrent connections, which propagate information across a sequence of words in an RNN. Conceptually, when an RNN is ``unrolled" it is equivalent to a feedforward neural network with an infinite
number of connected hidden layers stacked on top of one another. Because of this depth of hidden layers, it can potentially learn complex dependencies among words (Figure~\ref{fig:rnn}).

\begin{figure}[t]
\begin{center}
\includegraphics[width=0.75\textwidth]{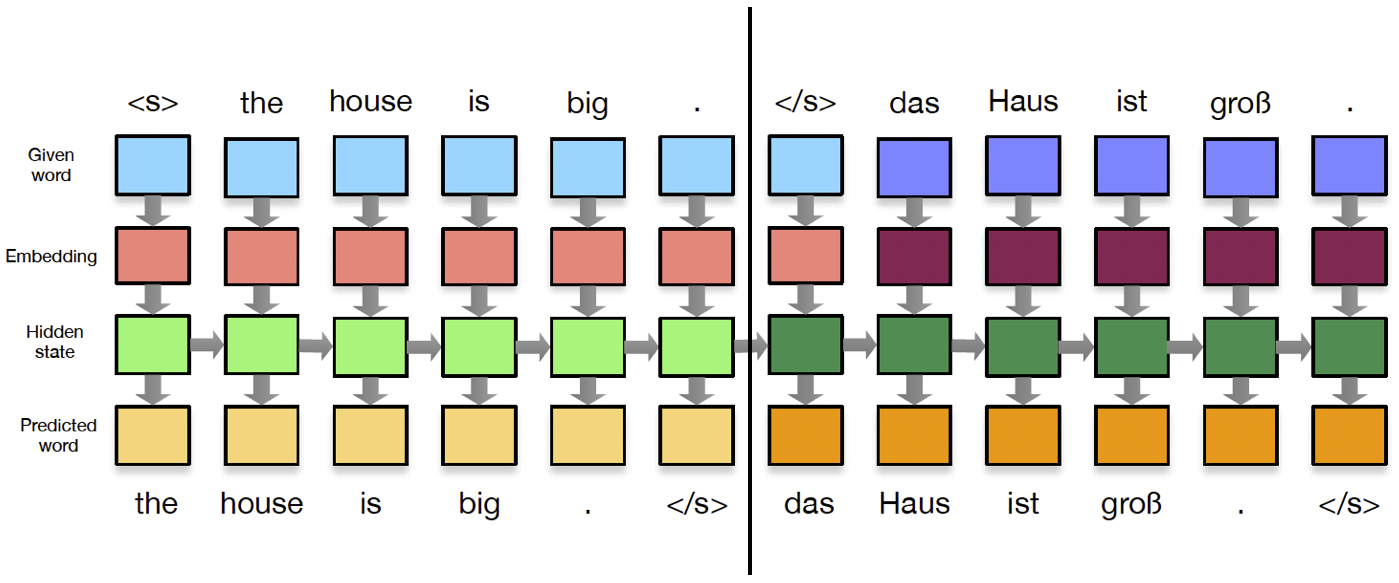}
\caption{The encoder-decoder model, also known as the sequence-to-sequence model (Seq2Seq). The light green boxes to the left of the vertical line make up the encoder RNN. The first dark box to the right of the vertical line encodes the entire source sentence. As we move from left-to-right, this source encoding is use to generate words in the target language.  (Image from \citep{koehn17})}
\label{fig:encdec}
\end{center}
\end{figure}

Fourth, \citeauthor{cho2014learning} (\citeyear{cho2014learning}) extended RNNs for machine translation by
creating the encoder-decoder model (also known as the sequence-to-sequence model (Seq2Seq)). This model concatenates two RNNs together, one that encodes the source language, and one that decodes the target language. In Figure~\ref{fig:encdec}, the light green boxes to the left of the vertical line make up the encoder RNN, which encodes the words in the source sentence. The first dark green box to the right of the vertical line encodes the entire source sentence. As we move from left-to-right, this source encoding is used to generate words in the target language. 
  
Fifth, a major advancement in NMT occurred through the use of a memory mechanism to align the source sequence positions to the target sequence state. \cite{cho2014properties} observe that Seq2Seq models deteriorate quickly as the input sequence length increases. This is because the decoder is forced to make a hard decision to predict a target word at every state. \cite{bahdanau&al15} propose an attention mechanism that allows the Seq2Seq model to focus on a set of positions from the source sentence to form a context vector that are most relevant to the current state in the target sequence. 
It uses the context vector from the attention mechanism to predict the current word. \cite{luong2015effective} extends the attention mechanism by introducing both global and local attention mechanisms. The global attention mechanism functions as described above by considering all positions in a source sentence; the local attention mechanism restricts its focus to the vicinity of source positions that best correspond to the target position that is to be predicted. Attentional Seq2Seq models are among the best performers on standard machine translation benchmarks. Hence, we employ one such model~\citep{luong2015effective} for our experiments in Section~\ref{sect:experiments}.  


Sixth, \cite{vaswani2017} create an NMT model called Transformer that dispenses with RNNs. RNNs  requires a time-consuming, left-to-right, word-by-word traversal of the entire input sentence in order to model the full span of a sentence. However, such a traversal is not parallelizable and severely slows down model training. 
By discarding RNNs, the Transformer model becomes highly parallelizable,  and it retains the ability to model the entire span of a sentence through the use of {\em self-attention}. In an attentional Seq2Seq model, the attention mechanism models the association between an output word with every input word. In self-attention, we compute the association between each input word and every other input word, thereby  disambiguating an input word using other input words as context.  Further, the Transformer uses {\em multi-head} self-attention, i.e., it applies self-attention in multiple representation spaces (e.g., one that captures the syntax of a language, and another that captures the morphology) to enrich the representation of a word. 
 The Transformer model is among the best performers on standard machine translation benchmarks, and we use it for our experiments in Section~\ref{sect:experiments}.

\begin{figure}
\centering
\begin{minipage}{.5\textwidth}
  \centering
  \includegraphics[width=.8\linewidth]{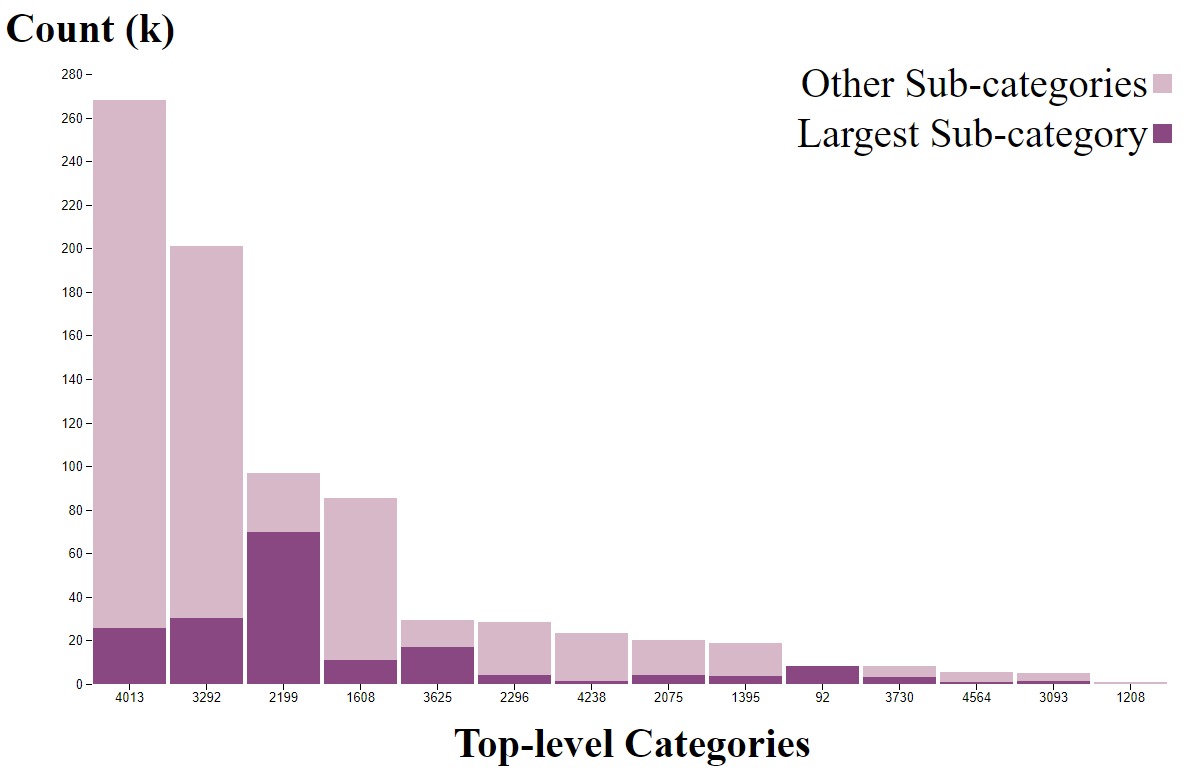}
  \captionof{figure}{Skewed product distribution in RDC.}
  \label{fig:datadistRDC}
\end{minipage}%
\begin{minipage}{.5\textwidth}
  \centering
  \includegraphics[width=.83\linewidth]{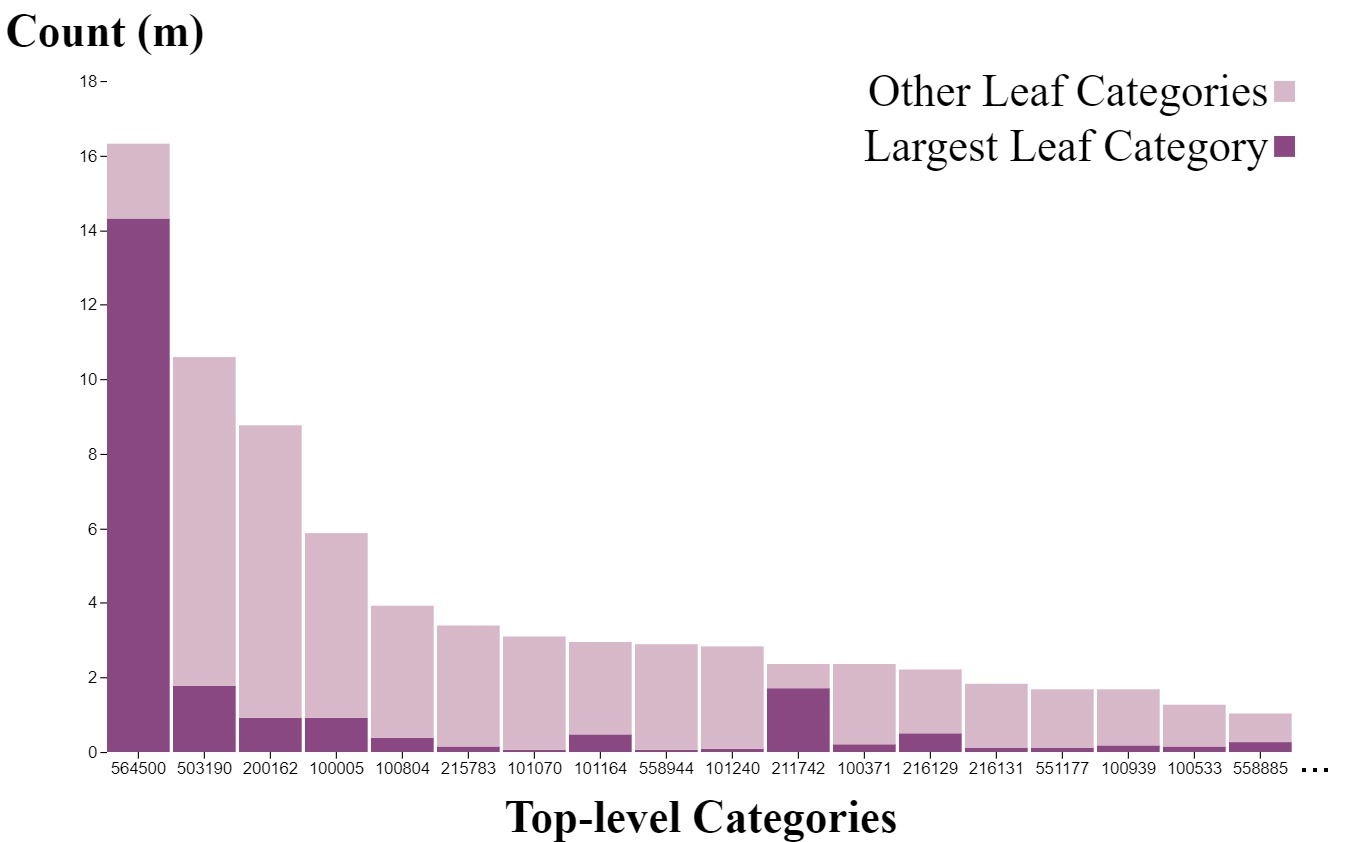}
  \captionof{figure}{Skewed product distribution in Ichiba.}
  \label{fig:datadistIchiba}
\end{minipage}
\end{figure}

\section{Experiments}
\label{sect:experiments}

\subsection{Datasets}

We used the following two e-commerce datasets for our experiments (Table~\ref{tab:data} provides a summary of their characteristics).

\begin{itemize}
\item {\bf Rakuten Data Challenge (RDC)}\footnote{\url{https://sigir-ecom.github.io/data-task.html}}. This dataset from \textit{www.rakuten.com} contains 800,000 product titles in English with their respective multi-level category labels. We lowercase all product titles and tokenize them with the Moses tokenizer \footnote{\url{https://github.com/moses-smt/mosesdecoder/blob/master/scripts/tokenizer/tokenizer.perl}}.

\item {\bf Rakuten Ichiba}\footnote{\url{https://rit.rakuten.co.jp/data_release/}}.  This dataset from \textit{www.rakuten.co.jp} consists of 280 million products listed by over 40,000 merchants and has 28,338 categories. We remove duplicate product listings and those in the `{\tt Others}' category that are erroneously assigned by merchants. After this, we are left with about 100 million Japanese product titles that are paired with their multi-level category labels. We tokenize the product titles with the MeCab Japanese segmenter~\citep{kudo2004applying}. 
\end{itemize}

\begin{table}[h]
\centering
    \begin{tabular}{l|r|r}
    ~                          			     & \textbf{Rakuten Data}    & ~          \\
    ~                                        & \textbf{Challenge (RDC)}       & \textbf{Ichiba}     \\ \hline
    \textbf{Language} 						 & English 					& Japanese \\
    \textbf{Source} 						 & \texttt{www.rakuten.com} & \texttt{www.rakuten.co.jp} \\
    \textbf{Data size }                   	 & 800,000       			& 100,436,907     \\
    \textbf{No. of First-Level Categories}	 & 14           		    & 35             \\
    \textbf{No. of Unique Categories}    	 & 3,008         			& 21,819         \\ \hline
    \textbf{Tokenization}    				 & Moses tokenizer 			& MeCab \\
    										 & + lowercase 			& \\
    \end{tabular}
\caption{Summary of Rakuten Data Challenge (RDC) and Ichiba Datasets}
\label{tab:data}
\end{table}

\noindent
In both datasets, the products are assigned to the leaf nodes of a taxonomy tree. For each product, we have its title (e.g., {\it ``Mix Pancake Waffle 24 OZ -Pack of 6"}) and its root-to-leaf path (e.g., {\tt Baking Supplies \ $\rightarrow$ \ Flour \& Dough \ $\rightarrow$ \ Pancake \& Waffle Mixes}). All the models in Section \ \ref{subsect:models} take the product title as input to predict its associated root-to-leaf path (we term this path the {\em label} of the product).

The distribution of products across categories in both datasets is skewed towards the most popular categories as is usually the case in e-commerce domains~\citep{he2016ups,xia17}. Figure~\ref{fig:datadistRDC} and \ref{fig:datadistIchiba} show the number of products in each category at the top-level of the taxonomy tree (each vertical bar reflects the number of products in that category). These figures show that a majority of products is assigned to a few categories, and the rest are spread across the remaining categories in a long tail. Further, the dark-colored portion of each bar represents the sub-category with the highest count within the top-level category, and the lighter tip of the bar represents all other sub-categories within that top-level category. As can be seen, the distribution within some top-level categories may also be skewed. 



We randomly split both the RDC and Ichiba datasets in a stratified manner into their respective training, validation and test sets in the proportion of 80/10/10.
The validation set was used to determine the early stopping criteria for our NMT models. 

\subsection{Models}
\label{subsect:models}


For our neural machine translation (NMT) models, 
we use the attentional Seq2Seq model of \cite{luong2015effective} and the Transformer model of \cite{vaswani2017} as implemented in the Fairseq toolkit (commit \texttt{5d99e13})\footnote{\url{https://github.com/pytorch/fairseq/tree/master/fairseq}} (see Section~\ref{sect:mt} for their descriptions). The hyperparameters of our models are given in Table~\ref{tb:hyperparams}. Note that the recurrent neural network (RNN) hidden layer size is specific to the attentional Seq2Seq model while the feedforward network (FFN) hidden layer and attention heads hyperparameters are only used in the Transformer model. We also ensemble the attentional Seq2Seq model and the Transformer model together by averaging their decoder outputs.

As mentioned, each product is associated with its title and root-to-leaf path. Our NMT models consider the title to be a sentence in a source language (English for RDC and Japanese for Ichiba), and translate it into a sequence of tokens corresponding to the nodes in a root-to-leaf path.


\begin{table}[!htpb]
\centering
    \begin{tabular}{l|rr}
    ~                                      & \textbf{Attentional Seq2Seq}  & \textbf{Transformer} \\ \hline
    \textbf{Input/Output Embedding Dimension} & 512  & 512         \\
    \textbf{RNN Hidden Layer Size}               & 1,024 & -           \\
    \textbf{FFN Hidden Layer Size}               & -    & 2,048        \\
    \textbf{Stacked Layers}                & 1    & 6           \\
    \textbf{Dropout}                       & 0.2  & 0.2         \\
    \textbf{Attention Heads}               & -    & 8           \\ \hline
    \textbf{No. of Parameters}             & 7,5435,103   & 99,105,792    \\
    \end{tabular}
\caption{Hyperparameters of attentional Seq2Seq and Transformer Models}
\label{tb:hyperparams}
\end{table}

We compare our machine translation models with a traditional classification-based system \texttt{CUDeep} \citep{cevahir2016large}  that achieved state-of-the-art performance on the Ichiba dataset (this model is described in Section~\ref{sect:related}).
 \texttt{CUDeep} trains a deep belief network using a stacked restricted Boltzmann machine architecture \citep{hinton2006reducing}, and learns an encoder that embeds a product title into a vector representation. Next, it trains a feedforward neural layer to map the vector representation to a predicted product category. Henceforth, we will term this model DBN. Aside from using deep belief networks, 
 \texttt{CUDeep} also uses K-nearest neighbors (KNN) \citep{cover1967nearest} to predict product categories by   mapping a product title to the 1-nearest neighbor's category that is seen in the training data. We also combined the outputs of the DBN and KNN models to form a DBN+KNN ensemble and averaged the probabilities of their category predictions to re-rank the predictions.
 
\subsection{Evaluation Metrics}

\cite{cevahir2016large} previously used $n$-best accuracy as the evaluation metric for the Ichiba dataset, and \cite{lin2018overview} applied the support weighted F-score as the metric to evaluate the Rakuten Data Challenge (RDC) dataset. To keep the comparisons consistent across datasets, we opted for the single metric of weighted F-score. This metric weighs the accuracy in each leaf node by its number of products, and is thus better suited for multi-class prediction in skewed datasets.
The multi-class F1 score is computed as follows:

\vspace{-3mm}
\begin{equation}
\begin{split}
TP_c  = |{ \hat{y}  }_c \cap{y}_c|  \;\;\;\;\;\;\;\;\;\;
 P_ c  = \frac{ TP }{ |{ \hat { y }  }_{ c }| }     \;\;\;\;\;\;\;\;\;\;    
 R_ c  = \frac{ TP }{ |{ y }_{ c }| }   \;\;\;\;\; \;\;\;\;\; \\[1.5pt]
 F_c   = \frac{ 2\cdot P_c \cdot R_c}{ P_c+R_c}  \;\;\;\;\;\;\;\;\;\; 
F_c^{ \ weighted} = \frac {1}{|C|} \sum _{ c\in C }^{  }{ TP_ c } \cdot F_c
 \end{split}
\label{eq:fscore}
\end{equation}

\noindent
where $C$ represents all possible categories/labels, $\hat y_c$ are the products that are labeled by the system as $c$, $y_c$ are products with $c$ as the true labels, and $TP_c, P_c, R_c, F_c, F_c^{weighted}$ are respectively the true positives, precision, recall, F-score, and weighted F-score for label $c$. Due to the highly skewed category distribution, we use the \textit{weighted} variant of the precision, recall and F-score~\footnote{\url{http://scikit-learn.org/stable/modules/generated/sklearn.metrics.precision_recall_fscore_support.html}}, where the scores across labels are summed and weighted by their true positive values. (NB: For weighted variants of precision, recall, and F-score, the F-score may not lie between precision and recall.)



\subsection{Results}
\label{subsect:result}

Table \ref{tb:results} reports the weighted precision (P), weighted recall (R), and weighted F-scores (F) on the test sets of the RDC and Ichiba datasets. These scores only deem a label (i.e., a predicted root-to-leaf path) to be correct if it is an exact match to the ground truth. As long as one node in the path is wrong (even when the leaf node is correct), the prediction is deemed wrong. Note that this penalizes our NMT models because they can predict novel root-to-leaf paths that do not exist in a taxonomy tree, and can thus arrive at the correct leaf nodes via multiple paths (and not only through the unique root-to-leaf path in the taxonomy). Even though {\tt CUDeep} also predicts a root-to-leaf path, that path is an existing one in the taxonomy tree and is uniquely determined by the leaf node. To allow for a consistent comparison with {\tt CUDeep}, we decided to determine correctness by the full root-to-leaf path. If we consider the correctness of the leaf nodes only, our results will surpass those shown below. 



\begin{table}[H]
    \begin{tabular}{l|l|ccc|ccc}
    ~                       & ~                         &  ~  & \textbf{RDC}     & ~     & ~ &  \textbf{Ichiba}     & ~     \\
    ~                       & ~                         & \textbf{P}     & \textbf{R}     & \textbf{F}     & \textbf{P}      & \textbf{R}     & \textbf{F}     \\ \hline
                  & \textbf{Deep Belief Net (DBN)}     & 72.19 & 74.72 & 72.86 & 78.09  & 78.29 & 77.52 \\
    \texttt{CUDeep}                & \textbf{K-Nearest Neighbors (KNN)} & 71.14 & 72.10 & 70.94 & 79.24  & 78.69 & 78.66 \\
                   & \textbf{DBN+KNN}                 & 73.46 & 75.57 & 73.85 & 82.65  & 82.27 & 82.05 \\ \hline
    ~                       & \textbf{Attentional Seq2Seq }        & 74.03 & 73.43 & 72.50 & 84.70   & 82.39 & 82.08 \\

Our       & \textbf{Transformer}       & 74.44 & 75.25 & 73.83 & 83.79  & 83.59 & \textbf{84.74} \\
NMT Models      & \textbf{Seq2Seq+Transformer}        & \textbf{75.22} & \textbf{75.65} & \textbf{74.19} & \textbf{85.08}      & \textbf{84.31}     & 84.26     \\
    \end{tabular}
\caption{Results of our NMT Systems vs \texttt{CUDeep} Classification Systems}
\label{tb:results}
\end{table}


\begin{table}[H]
\begin{center}
    \begin{tabular}{l|ccc|ccc}
    &	\multicolumn{6}{c}{RDC} \\
     & \multicolumn{3}{c}{\textbf{p5}} & \multicolumn{3}{c}{\textbf{p95}} \\
& \textbf{P} &	\textbf{R} &	\textbf{F} & \textbf{P} &	\textbf{R} &	\textbf{F} \\ \hline
    \textbf{DBN+KNN} & 73.08 & 75.17 & 73.32 & 74.11 & 75.98 & 74.22 \\
    \textbf{Seq2Seq+Transformer} & 74.81 & 75.23 & 73.68 & 75.76 & 76.05 &74.57 \\
    \end{tabular}
\caption{Confidence Interval of 1000 iterations of Bootstrap Resampling on the Best Performing Models on RDC dataset}
\label{tb:bootstrap-results-rdc}
\end{center}
\end{table}

\begin{table}[H]
\begin{center}
    \begin{tabular}{l|ccc|ccc}
    &	\multicolumn{6}{c}{Ichiba} \\
     & \multicolumn{3}{c}{\textbf{p5}} & \multicolumn{3}{c}{\textbf{p95}} \\
 & \textbf{P} &	\textbf{R} &	\textbf{F} & \textbf{P} &	\textbf{R} &	\textbf{F} \\ \hline
    \textbf{DBN+KNN} & 82.64  & 82.24 & 82.00 & 82.71 & 82.30 &  82.07 \\
    \textbf{Seq2Seq+Transformer} & 85.07 & 84.28 & 84.22 & 85.13 & 84.35 & 84.28\\
    \end{tabular}
\caption{Confidence Interval of 1000 iterations of Bootstrap Resampling on the Best Performing Models on Ichiba dataset}
\label{tb:bootstrap-results-ichiba}
\end{center}
\end{table}


From Table~\ref{tb:results}, we see that our Transformer model outperforms both \texttt{CUDeep} single models (DBN and KNN) on both datasets (weighted F-scores of 73.83 and 84.74 on the RDC and Ichiba test sets respectively). The Transformer also outperforms the DBN+KNN ensemble on the Ichiba dataset, and is competitive on the RDC dataset. Our attentional Seq2Seq model has mixed results on RDC, but outperforms all {\tt CUDeep} models for all metrics on the larger Ichiba dataset. Our Seq2Seq+Transformer ensemble is the best performer across the board. It is better than both our single models and all {\tt CUDeep} models. The only exception is that the weighted F-score of our Seq2Seq+Transformer model is marginally lower than that of our Transformer model.


To further confirm that our system outperforms \texttt{CUDeep}, we conducted 1000 iterations of bootstrap resampling on the best performing model in each system to find out the 95\% confidence interval of their performance scores. From Tables~\ref{tb:bootstrap-results-rdc} and \ref{tb:bootstrap-results-ichiba}, we see can conclude that our attentional Seq2Seq+Transformer ensemble surpasses {\tt CUDeep}'s DBN+KNN ensemble.


\begin{table}[H]
\begin{center}
    \begin{tabular}{l|cccc}
      & \textbf{80-10-10} & \textbf{60-10-30} & \textbf{40-10-50} & \textbf{20-10-70} \\ \hline
    \textbf{DBN+KNN}             & 73.85        & 74.08        & 71.24        & 61.27        \\
    \textbf{Seq2Seq+Transformer} & 74.19        & 74.94        & 73.77        & 69.58        \\
    \end{tabular}
\caption{Effects of Data Size with Respect to Systems' Weighted F-score}
\label{tb:incr-trg-splits}
\end{center}
\end{table}

\vspace{-5mm}

We investigated the effect of training data size on the performances of the systems. Table~\ref{tb:incr-trg-splits} presents the F-scores of the ensembled systems with respect to various train-validation-test 
sizes. For instance, ``60-10-30'' indicates that a model was trained, validated and tested on 60\%, 10\% and 30\%  of the data respectively. 

We note that the 60-10-30 split has higher F-scores than the 80-10-10 split for both ensembled systems. This is due to the random split of the 80-10-10 data giving its test set a higher proportion of classes with one instance (i.e., these classes do not appear in the training set). The instances of such classes are impossible to correctly predict for both systems. From Table~\ref{tb:incr-trg-splits}, we see that our machine-translation-based Seq2Seq+Transformer ensemble is consistently more robust to reductions in data sizes than the DBN+KNN ensemble. Even with only 20\% of the training data, the performance of our Seq2Seq+Transformer ensemble does not degrade as much as that of the DBN+KNN model. Further, our Seq2Seq+Transformer ensemble consistently outperforms the DBN+KNN model across data sizes.

\section{Analysis}
\label{sect:analysis}

\begin{table}[!htbp]
\centering
    \begin{tabular}{l|rr}
    ~                 & \textbf{RDC}            & \textbf{Ichiba}  \\ \hline
    \textbf{RNN}               & 106            & 5,183   \\
    \textbf{Transformer}       & 76             & 113,287 \\
    \textbf{RNN + Transformer} & 124            & 48,455       \\
    \end{tabular}
\caption{Count of Full-Path Categories Created}
\label{tb:new-cat}
\end{table}

As discussed in previous sections, our NMT models generate root-to-leaf paths based on the vocabulary of categories. This generation allows new paths to be created based on product titles. Although such system-created paths utilize {\it existing} nodes in a product taxonomy tree, the paths (which are permutations of nodes) need not pre-exist in the tree. When the paths are added to the tree to form new edges between nodes, they  transform the tree into a DAG, which offers a richer representation of the products.

We present the count of novel categorization paths created by each of our models in Table~\ref{tb:new-cat}. In this section, we qualitatively analyze some notable examples of created paths in the English-language RDC dataset.



The product `\textit{Hal Leonard Neil Young-Rust Never Sleeps Guitar Songbook}' has its ground truth root-to-leaf path as {\tt Home \& Outdoor > Hobbies > Musical Instruments > Misc Accessories > Sheet Music}. Our Transformer model's predicted path is identical to the ground truth except that it omits {\it Musical Instruments} from the path. This is intuitively correct because the product is a songbook, which does not belong to the {\it Musical Instruments} sub-category. This example suggests that our NMT models can prune and restructure the taxonomy tree to more accurately describe products.



Another notable example is the product \textit{Epson WorkForce Pro WP-4023 Inkjet Printer C11CB30 231 Compatible 10ft White}, which has the ground truth category of {\tt Electronics>...>Printers}. Our NMT model predicts the root-to-leaf path as {\tt Office Supplies>...>Printers}, which is intuitively correct because printers constitute general office supplies. This suggests that our NMT models can enrich the representation of products. 

Our system-created paths are not constrained by the existing hierarchical ordering of nodes in a taxonomy tree (e.g., it can place a leaf-category node at its start and a top-level-category node at its end). However, we observe that the paths created in our experiments all begin with top-level-category nodes and end with leaf nodes. This is because our machine translation models have successfully learned from their training data the strong bias of top-level-category nodes to appear first and leaf nodes to appear last. Beyond that, the paths conform less to the structure of the taxonomy tree, with some spanning across branches, and moving from lower-level categories to higher-level ones.

\section{Conclusion \& Future Work}
\label{sect:conclusion}

Product categorization is an important problem for e-commerce companies. By changing the framing of the problem from the traditional one of classification to one of machine translation, we show that state-of-the-art machine translation (MT) models surpass previous classification approaches in categorizing products in two large real-world e-commerce datasets. 

Besides enhancing the performance of product categorization, our NMT models also create novel root-to-leaf category paths. These novel paths can help to adapt a product taxonomy to changes in product listings. They also suggest ways to restructure the product taxonomy so that the category paths better accommodate a user's multiple conceptualizations of an product.

Future work includes: 
crowdsourcing the evaluation of novel root-to-leaf paths,
experiments with more MT models and classification models,  
automatic induction of the product taxonomy from data, etc.

\vspace{2mm}
\noindent
{\bf Acknowledgements.} This work was done during Maggie Li's internship at the Rakuten Institute of Technology Singapore (RIT-SG). We thank RIT for the collaboration, support and computation resources for our experiments. We also thank Ali Cevahir for his advice on the \texttt{CUDeep}-related experiments. 

%



\bibliographystyle{apalike}
\bibliography{references.bib,refs.bib,mt.bib,exp.bib}

\end{document}